# Are we still able to recognize pearls? Machine-driven peer review and the risk to creativity: An explainable RAG-XAI detection framework with markers extraction

Alin-Gabriel Văduva, Simona-Vasilica Oprea, Adela Bâra
Bucharest University of Economic Studies, Department of Economic Informatics and Cybernetics, no. 6 Piaţa Romană, Bucharest, Romania
*Corresponding author: simona.oprea@csie.ase.ro

**Abstract:** The integration of large language models (LLMs) into peer review raises a concern beyond authorship and detection: the potential cascading automation of the entire editorial process. As reviews become partially or fully machine-generated, it becomes plausible that editorial decisions may also be delegated to algorithmic systems, leading to a fully automated evaluation pipeline. They risk reshaping the criteria by which scientific work is assessed. This paper argues that machine-driven assessment may systematically favor standardized, pattern-conforming research while penalizing unconventional and paradigm-shifting ideas that require contextual human judgment. We consider that this shift could lead to epistemic homogenization, where researchers are implicitly incentivized to optimize their work for algorithmic approval rather than genuine discovery. To address this risk, we introduce an explainable framework (RAG-XAI) for assessing review quality and detecting automated patterns using markers LLM extractor, aiming to preserve transparency, accountability and creativity in science. The proposed framework achieves near-perfect detection performance, with XGBoost, Random Forest and LightGBM reaching 99.61% accuracy, AUC-ROC above 0.999 and F1-scores of 0.9925 on the test set, while maintaining extremely low false positive rates (<0.23%) and false negative rates (~0.8%). In contrast, the logistic regression baseline performs substantially worse (89.97% accuracy, F1-score 0.8314). Feature importance and SHAP analyses identify *absence of personal signals* and *repetition patterns* as the dominant predictors. Additionally, the RAG component achieves 90.5% top-1 retrieval accuracy, with strong same-class clustering in the embedding space, further supporting the reliability of the framework's outputs.

**Keywords:** peer-review; AI manuscript assessment; detection framework; markers; editor report tool; LLM-feature extractor; explainable AI; retrieval-augmented generation

## 1. Introduction

The proliferation of large language models (LLMs) has introduced new challenges to the integrity of the review process. While these models can assist researchers in drafting text or summarizing literature, they can also be used to generate entire peer reviews with minimal human oversight. As a result, concerns have emerged regarding the authenticity, reliability, ethics and accountability of reviews submitted to academic journals [1]. Models based on transformer architecture have significantly improved the ability of machines to produce coherent and contextually relevant text. They generate structured feedback that resembles conventional peer reviews, including summaries, lists of strengths and weaknesses, and detailed methodological critiques, creating the illusion that reviews may be partially or entirely AI-generated, raising important questions for editors, publishers and research communities regarding transparency, accountability and the preservation of rigorous scholarly evaluation [2], [3].

The current peer-review process could be increasingly marked by a disconnect between authors' expectations and the quality of feedback received, with long delays often culminating in superficial, generic or misaligned reviews that fail to engage with the manuscript's actual contributions, frequently citing standard criticisms (e.g., missing ablation studies) regardless of relevance. This issue is exacerbated by the tacit use of LLMs by reviewers, which can produce fluent but shallow and templated evaluations, leading to homogenized and less meaningful assessments [4]. Simultaneously, a new form of adversarial behavior is emerging, where authors attempt to manipulate AI-assisted reviewers by embedding hidden prompts (e.g., small-sized white-font text) within manuscripts to influence evaluation outcomes [5].

A further and more fundamental concern arises from the potential cascading automation of the entire editorial process. If peer reviews are increasingly generated/assisted by machines, it becomes plausible that editorial decisions, traditionally grounded in expert judgment, may also be delegated to algorithmic systems. In such a scenario, both evaluation and decision-making would be performed by machines, raising questions about the future of creativity. Scientific progress has depended on the recognition of unconventional and paradigm-shifting ideas, which require human judgment and contextual understanding. An evaluation ecosystem dominated by machine-driven assessments may favor standardized, pattern-conforming research while penalizing originality and deviation from learned norms, leading to a form of epistemic homogenization, where researchers are implicitly incentivized to produce work optimized for algorithmic approval rather than genuine scientific advancement. Consequently, the perception that "machines judge science" may erode researchers' intrinsic motivation for innovative or exploratory work, ultimately posing a long-term threat to the diversity, creativity and vitality of scientific inquiry [6], [7].

Thus, detecting AI-generated text has become a challenge. Existing approaches typically rely on statistical signals, stylometric features or classifier-based detection models trained to differentiate human and machine-generated text [8], [9]. However, these approaches face several limitations. First, modern LLM outputs increasingly mimic human writing styles, reducing the reliability of purely statistical detectors. Second, many detectors operate as opaque classifiers that provide a binary judgment without explaining which linguistic or structural features influenced the decision. Third, peer-review texts differ from general prose in their structure and purpose, often containing domain-specific terminology, structured feedback patterns and evaluation criteria. As a result, generic AI-detection models may not perform reliably when applied to the peer-review domain [10].

The proposed approach focuses on classification and also on explainable detection. Instead of simply predicting whether a review is likely to be AI-generated, the system highlights specific linguistic and structural markers that may indicate automated generation. These markers may include patterns such as highly regularized rhetorical structures, unusually balanced argumentation, absence of line-level comments, stylistic homogeneity or repeated templated phrasing across sections [11]. By grounding the detection process in retrieved evidence from curated corpora of peer reviews, the framework can provide interpretable explanations that support editorial decision-making.

The motivation for such a framework is twofold. First, authors increasingly report concerns about the use of generative AI in the peer-review process, particularly in fast-moving fields such as AI and computer science. Second, the availability of AI-assisted writing tools raises questions about where the boundary lies between legitimate assistance and fully automated review generation. A transparent detection system could therefore help editors identify suspicious reviews, encourage disclosure of AI assistance and maintain trust in the peer-review system.

In this work, we propose an LLM-feature extractor for classifiers with RAG-XAI framework for identifying potential AI-generated peer reviews and for extracting interpretable indicators of automated text generation. The framework integrates a curated dataset of human and AI-generated reviews, retrieval-based similarity analysis (RAG) and a marker-detection layer designed to capture structural and stylistic signals specific to the peer-review genre. The editor benefits from obtaining an AI-generated probability, explainable markers and a retrieval system that will give similar reviews from RAG knowledge base. Our study aims to contribute both methodological insights and practical tools[1] for maintaining integrity in scholarly evaluation.

## 2. Literature review

The integration of LLMs into the peer-review process has generated increasing scholarly attention, reflecting both their transformative potential and the risks they pose to scientific integrity. Early conceptual discussions emphasize that AI is no longer confined to auxiliary research tasks but is progressively influencing all stages of the publication pipeline, including manuscript evaluation and editorial decision-making [12]. A significant enabler of empirical investigation is the availability of large-scale datasets. The GenReview dataset [13] represents a major advancement by providing over 81,000 LLM-generated reviews linked to real ICLR submissions, enabling systematic analysis of bias, instruction-following and alignment with editorial decisions. The study finds that LLMs exhibit measurable reviewing biases, inconsistently adhere to evaluation criteria and align with acceptance decisions only under specific conditions, highlighting both their capabilities and limitations. Parallel research has focused on quantifying the prevalence and influence of AI-assisted peer reviews. Empirical evidence from ICLR 2024 [3] shows that at least 15.8% of reviews involved AI assistance, with such reviews tending to assign higher scores and increasing the likelihood of acceptance for borderline submissions. Complementary corpus-level analyses [14] estimate that 6.5%–16.9% of peer review text across major AI conferences is AI-generated, while temporal studies [15] reveal a sharp post-2022 increase, reaching approximately 20% of reviews in some venues. Linguistic analyses [16] further identify characteristic stylistic patterns of AI-generated reviews, including verbosity, generic phrasing and lack of technical specificity.

---

[1] https://review-detector-analysis.streamlit.app/

A substantial body of work addresses the detection of AI-generated peer reviews, which remains a challenging problem. Early studies [17] demonstrate that existing detection tools struggle to distinguish LLM-generated reviews from human-written ones without producing high false positive rates. To improve detection, recent approaches incorporate more sophisticated representations. For instance, syntactic-semantic frameworks [18] combine discourse-level and semantic features to achieve high detection accuracy, while MixRevDetect [19] focuses on identifying AI-generated segments within hybrid reviews. Alternative approaches include watermarking-based detection [20], which embeds detectable signals into generated text. Despite these advances, empirical findings consistently highlight that detection remains unreliable, particularly in cases of partial AI assistance/well-edited outputs [8].

Beyond detection, several studies evaluate the quality and reliability of LLM-generated reviews. Experimental evidence [21] shows that LLM-generated reviews can appear convincing but lack depth in methodological critique and may introduce irrelevant or manipulative suggestions. Similarly, controlled evaluations of ChatGPT [22] reveal limitations in identifying critical inconsistencies and ensuring accurate referencing. Robustness analyses [23] further demonstrate that LLM-based reviewers are vulnerable to adversarial manipulation, raising concerns about their reliability in high-stakes evaluation settings, aligning with broader observations that automated tools lack the contextual understanding required to assess scientific validity and methodological soundness [24].

Another critical dimension concerns bias and fairness. Controlled experiments [25] identify systematic affiliation bias favoring highly ranked institutions, as well as subtle gender biases in LLM-generated reviews. Diagnostic frameworks [26] further reveal structural vulnerabilities, including misinterpretation of methodological flaws and disproportionate sensitivity to certain review signals, suggesting that LLMs may reproduce or amplify existing inequities in scholarly evaluation.

In response to these limitations, a growing body of research explores the integration of LLMs into peer-review systems. Simulation frameworks such as AGENTREVIEW [27] model peer review as a multi-agent system and demonstrate that reviewer biases alone can lead to substantial variability in editorial decisions. Similarly, automated systems such as AutoRev [28] leverage retrieval-augmented generation to produce structured and high-quality feedback. Comprehensive surveys [29], [30] synthesize these developments, identifying a wide range of LLM applications in peer review, including reviewer assistance, automated feedback generation, bias detection and editorial decision support, while emphasizing persistent limitations in reasoning, domain expertise and interpretability.

Empirical studies also examine the interaction between human reviewers and AI tools. Experimental evidence [31] shows that LLM assistance reduces reviewer workload but does not significantly improve review quality, while qualitative research [32] highlights concerns about transparency, bias and loss of autonomy. Survey-based studies [33], [34] indicate that researchers generally support AI-assisted peer review under human supervision, reinforcing the view that LLMs are best positioned as complementary tools rather than replacements for human judgment.

Further, several contributions address the ethical and governance implications of AI in peer review. Case-based evidence [4] illustrates the practical challenges of identifying and addressing suspected AI-generated reviews, including the absence of clear editorial protocols. Broader discussions [35], [36] emphasize the need for transparent policies, disclosure requirements and accountability mechanisms. Additional perspectives [37] highlight risks related to authorship, data integrity and misuse of AI-generated content, proposing mechanisms such as post-publication peer review to enhance accountability. A comparative perspective of the works on detecting AI-generated reviews is provided in Table 1.

Table 1. Comparison of several studies on detecting AI-generated reviews

| Ref | Objective | Methods / Tools | Results / Main Findings |
|---|---|---|---|
| [3] | Estimate prevalence of AI-assisted peer reviews and assess impact | Closed-source & open-source LLM detectors; statistical comparison of review scores | ≥15.8% reviews AI-assisted; AI reviews give higher scores and increase acceptance probability (+4.9 pp) |
| [8] | Identify practical indicators of AI-generated peer reviews | Editorial experience; qualitative analysis; detection heuristics | Detection relies on patterns (tone, structure); need for guidelines |
| [13] | Assess detectability of LLM-generated reviews using large dataset (GenReview) | Synthetic dataset (81K reviews); classification experiments | LLM reviews are currently detectable, but detection depends on dataset and conditions |

| [14] | Estimate proportion of AI-generated text in peer-review corpora | Maximum likelihood estimation; corpus-level statistical modeling | 6.5–16.9% of review text AI-generated; detection works better at aggregate level, not individual |
| --- | --- | --- | --- |
| [15] | Measure temporal growth of AI-generated peer reviews | Detection model trained on historical data | Near-zero pre-2022→~20% (ICLR) and ~12% (Nature Comms) in 2025 |
| [16] | Identify linguistic signals of AI-generated peer reviews | Large-scale linguistic/statistical analysis (~50K reviews) | 7–17% AI-generated; detectable patterns: verbosity, generic, lack of specificity |
| [17] | Evaluate performance of existing AI detectors on peer reviews | Benchmarking GPT-4o vs human reviews; detector comparison | Existing detectors fail frequently; high false positives; new method improves detection but problem remains hard |
| [18] | Improve detection accuracy using discourse-aware features | Syntactic-semantic framework; RSTGraphNet; entity consistency | High performance (F1≈0.9586); strong generalization and low false positives |
| [19] | Detect AI-generated segments within hybrid peer reviews | MixRevDetect model; fine-grained classification; dataset+ML | F1≈88.86%; outperforms traditional detectors; handles partial AI usage |
| [20] | Develop robust detection method via watermarking | Indirect prompt injection; statistical hypothesis testing; watermark embedding | High detection accuracy with controlled error rates; more reliable than traditional detectors |
| [21] | Evaluate detectability and risks of LLM-generated peer reviews in biomedical context | AI detectors (ZeroGPT, GPTZero); expert validation | 82.8% of AI reviews misclassified as human→detection tools largely fail |

## 3. Methodology and data

This section formalizes the proposed framework for detecting AI-generated peer reviews. We first define the problem and notation, then describe the dataset construction, introduce the eight-marker taxonomy and the LLM-based extraction procedure, present the multi-classifier training protocol, detail the SHAP explainability layer, formalize the retrieval-augmented generation module and describe the automated editor report generation.

### 3.1. Problem formulation

Let $R=\{r_1, r_2, …, r_n\}$ denote a corpus of $N$ peer review texts. Each review $r_i$ is associated with a ground-truth label $y_i\in\{0, 1\}$, where $y_i=0$ indicates a human-authored review and $y_i=1$ indicates an AI-generated review. The objective is to learn a classification function $f$ that maps a review to its predicted label:

$$f : R \rightarrow \{0, 1\} \qquad (1)$$

The framework first extracts a fixed-dimensional feature representation. Let $\varphi$ denote a feature extraction function that maps a review text to a vector of $m$ marker scores:

$$\varphi(r_i) = x_i = (x_{i1}, x_{i2}, …, x_{im}) \in [0, 1]^m \qquad (2)$$

where $m=8$ corresponds to the number of linguistic markers defined in our taxonomy. Each component $x_{ij}\in[0, 1]$ quantifies the degree to which marker $j$ is present in review $r_i$. The classifier then operates on this extracted feature space:

$$\hat{y}_i = f(x_i) = f(\varphi(r_i)) \qquad (3)$$

By decoupling the feature representation from the classifier, the framework gains interpretability: the eight marker scores are semantically meaningful to domain experts (journal editors), enabling human-in-the-loop decision-making.

### 3.2. Data collection and preparation

The experimental dataset comprises N=7,772 peer reviews drawn from both human and AI-generated sources. The construction follows a multi-source strategy designed to test classifier robustness against varying degrees of generation sophistication.

*Human reviews (n=5,772).* Human-authored reviews are extracted from the PeerRead[2] corpus [38], a publicly available dataset of scientific peer reviews from three major computer science venues: ICLR 2017 (n=5,458), ACL 2017 (n=275) and CoNLL 2016 (n=39). The reviews are parsed from the original JSON files, extracting the *comments* field from each review object. Reviews shorter than 50 characters are excluded to ensure substantive content.

The AI-generated reviews (n=2,000) are produced using two distinct strategies, yielding reviews of varying detectability:

---

[2] https://github.com/allenai/PeerRead

- *Neutral AI reviews (n=1,000).* Generated using GPT-4o through the Gen-Review framework[3] with a standard prompt instructing the model to produce a complete peer review including summary, strengths, weaknesses, questions for authors and overall recommendation. These reviews were generated by the Gen-Review framework with a temperature of 0.7, a maximum of 1,500 tokens and no explicit instruction to mimic human style. As expected, they exhibit strong AI-typical patterns: rigid sectional structure, formulaic criticism and absence of personal voice markers;
- *Adversarial AI reviews (n=1,000).* Designed to be deliberately harder to detect, these reviews simulate a realistic evasion scenario where an actor actively attempts to disguise AI-generated content as human-written (as in Appendix A, Prompt A). The adversarial subset is split equally between two state-of-the-art LLM providers: Claude Sonnet 4.6 via Amazon Bedrock (n=500) and GPT 5.4 (n=500). The adversarial prompts explicitly instructed the models to: (a) avoid formulaic section headers, (b) include personal voice markers such as hedging and self-correction, (c) vary sentence length and structure, (d) reference specific pages, figures and line numbers and (e) adopt an informal, conversational tone where appropriate.

Thus, to evaluate the robustness of the detection framework against deliberate evasion, we generate the 1,000 adversarial reviews using a prompt specifically designed to suppress the linguistic markers targeted by the taxonomy. The adversarial prompt instructs the LLM to adopt the persona of an experienced human peer reviewer and provides explicit stylistic directives that directly counteract each marker category.

The prompt specifies multiple style requirements, including: (i) avoiding markdown headers and templated structure (targeting *standardized_structure*), (ii) including personal voice markers such as "I think" and "after reading this twice" (targeting *absence_personal_signals*), (iii) referencing specific page numbers, figure numbers and equation numbers (targeting *conceptual_feedback*), (iv) mixing formal and informal language with varied sentence lengths (targeting *linguistic_homogeneity*), (v) expressing genuine uncertainty (targeting *excessive_balance*), (vi) avoiding balanced Strengths/Weaknesses lists (targeting *predictable_criticism*) and (vii) including minor imperfections such as tangents and self-corrections (targeting *repetition_patterns*). Each review is generated based on the title, abstract, and content of a real ICLR submission drawn from the Gen-Review corpus, with temperature set to 0.9 to maximize stylistic diversity. The identical prompt is used for both Claude Sonnet 4.6 and GPT 5.4, enabling a controlled comparison of evasion effectiveness across LLM architectures.

Table 2 summarizes the composition of the final dataset. The deliberate inclusion of adversarial reviews is a methodological contribution of this work, as most existing AI-text detection studies evaluate only against naively generated text, which overstates the detection performance.

Table 2. Dataset composition

| Source | Type | Generator | N | % of total |
|---|---|---|---|---|
| *PeerRead-ICLR 2017* | Human | - | 5,458 | 70.2% |
| *PeerRead-ACL 2017* | Human | - | 275 | 3.5% |
| *PeerRead-CoNLL 2016* | Human | - | 39 | 0.5% |
| *Gen-Review (neutral)* | AI | GPT-4o | 1,000 | 12.9% |
| *Adversarial* | AI | Claude Sonnet 4.6 | 500 | 6.4% |
| *Adversarial* | AI | GPT 5.4 | 500 | 6.4% |
| *Total* | - | - | 7,772 | 100% |

The combined dataset is shuffled with a fixed random seed and stored as a CSV file[4].

*3.3. LLM-based feature extraction*

Rather than relying on traditional surface-level linguistic features (e.g., word frequency distributions, readability indices or perplexity scores), we introduce a taxonomy of eight higher-order textual markers that capture the semantic, pragmatic and behavioral characteristics of AI-generated writing. The markers are organized into four categories: structural, argumentative, linguistic and behavioral, as detailed in Table 3.

Table 3. Taxonomy of AI-generation markers

| Category | Marker | Description |
|---|---|---|

---

[3] https://arxiv.org/pdf/2510.21192
[4] https://github.com/alingabriel743/review-detector/blob/main/data/dataset.csv

| | | |
|---|---|---|
| *Structural* | Standardized Structure ($x_1$) | Degree of rigid template adherence (e.g., labeled Summary / Strengths / Weaknesses sections) |
| *Argumentative* | Predictable Criticism ($x_2$) | Reliance on generic methodological critiques (e.g., "needs ablation study", "stronger baselines") |
| *Argumentative* | Excessive Balance ($x_3$) | Overly diplomatic tone that systematically pairs every criticism with positive framing |
| *Linguistic* | Linguistic Homogeneity ($x_4$) | Uniformity of grammar, sentence length and tonal register throughout the review |
| *Linguistic* | Generic Domain Language ($x_5$) | Use of broad academic phrases ("novel approach", "significant contribution") vs. precise technical terms |
| *Behavioral* | Conceptual Feedback ($x_6$) | High-level feedback without specific references to lines, pages, figures or tables |
| *Behavioral* | Absence of Personal Signals ($x_7$) | Lack of subjective voice markers ("I think", "I struggled with", hedging, self-correction) |
| *Behavioral* | Repetition Patterns ($x_8$) | Presence of templated or repeated phrasing patterns across sections of the review |

This taxonomy is grounded in empirical observations from the AI-generated text detection literature. LLMs exhibit systematic biases at levels deeper than simple lexical statistics. The *Absence of Personal Signals* marker ($x_7$) captures the well-documented tendency of LLMs to produce impersonal, authoritative prose that lacks the hedging, self-correction and subjective expressions characteristic of human reviewers. The *Linguistic Homogeneity* marker ($x_4$) reflects the lower variance in sentence structure that arises from the autoregressive generation process.

Each review $r_i$ is scored by prompting Claude Sonnet 4.6 with a structured zero-temperature prompt (Appendix A, Prompt B). The prompt provides the model with an explicit rubric defining each of the eight markers, followed by the review text (truncated to 3,000 characters to respect context window limits). The model is instructed to return only valid JSON containing the eight scores. Temperature is set to T=0.0 to ensure deterministic, reproducible scoring. To maximize throughput, extraction is parallelized using a thread pool of 10 concurrent workers, with each thread instantiating its own Bedrock client for thread safety. Progress is checkpointed to disk every 50 reviews to prevent data loss during long extraction runs.

A design choice in the LLM-as-feature-extractor paradigm is the framing of the extraction prompt (Prompt B). Early experiments revealed that prompts framing the LLM as "an expert in detecting AI-generated reviews" produced highly polarized scores, near-ceiling values (0.85–0.95) for AI-generated text and near-floor values (0.10–0.20) for human text, suggesting a self-recognition bias in which the LLM identifies its own outputs. To mitigate this effect, we adopted a neutral framing that instructs the model to act as a linguistic analyst with no knowledge of the detection task.

Each property is accompanied by a brief operational definition. For example, *standardized_structure* is defined as "How rigidly does the text follow a templated structure with clearly labeled sections (e.g., Summary, Strengths, Weaknesses)?", while *absence_personal_signals* is defined as "How absent are personal voice markers (e.g., 'I think', 'I found', 'in my experience', expressions of uncertainty)?". The model is instructed to return only a JSON object containing the eight numerical scores, with no justifications or additional text, to minimize parsing failures and ensure consistency across the corpus.

For environments where LLM API access is unavailable, a deterministic rule-based extractor is provided. Each marker is computed using handcrafted heuristics:

- **Standardized structure ($x_1$):** Computed as $\min(h/5, 1.0)$, where $h$ is the count of section header keywords (e.g., "summary", "strengths", "weaknesses") detected via regular expression matching;
- **Predictable criticism ($x_2$):** Computed as $\min(c/4, 1.0)$, where $c$ is the count of generic critique phrases (e.g., "ablation study", "stronger baseline", "additional dataset") found via case-insensitive substring matching against a curated lexicon of 11 formulaic expressions;

- **Excessive balance ($x_3$):** Computed as min($d$/3, 1.0), where $d$ counts diplomatic phrases (e.g., "well-written overall", "could be strengthened");
- **Linguistic homogeneity ($x_4$):** Computed as max(0, 1−$CV$), where $CV=\sigma(L)/\mu(L)$ is the coefficient of variation of sentence lengths L={$l_1, l_2, …, l_k$}. Lower variation yields a higher score;
- **Generic domain language ($x_5$):** Computed as min($g$/4, 1.0), where $g$ counts generic academic phrases (e.g., "state-of-the-art", "comprehensive framework");
- **Conceptual feedback ($x_6$):** Computed as max(0, 1−$p$/5), where $p$ is the count of specific line, page, figure or table references detected via regular expression. Reviews with zero such references receive a score of 1.0;
- **Absence of personal signals ($x_7$):** Computed as max(0, 1−$s$/3), where $s$ counts personal voice markers (e.g., "I think", "I may be wrong", "after re-reading") from a curated list of 14 expressions;
- **Repetition patterns ($x_8$):** Computed as min(3×(1−$u$), 1.0), where $u$=|unique trigrams|/|total trigrams| is the trigram uniqueness ratio. The factor of 3 amplifies the signal, as even small deviations from uniqueness are indicative.

The coefficient of variation formula for the *Linguistic homogeneity* marker is given by:

$$x_4 = max\left(0,\ 1 - \frac{\sigma(L)}{\mu(L) + \varepsilon}\right) \quad (4)$$

where $\sigma(L)$ and $\mu(L)$ denote the standard deviation and mean of sentence lengths respectively and $\varepsilon=10^{-9}$ is a small constant to avoid division by zero.

*3.4. Multi-classifier training and evaluation*

The eight marker scores $x_i=(x_{i1}, …, x_{i8})$ constitute the feature vector for binary classification. Four classifiers of increasing complexity are trained and compared to assess whether the marker features provide sufficient discriminative power across different model families.

The dataset of N=7,772 labelled reviews is split into training (70%) and test (30%) sets using stratified random sampling to preserve the class distribution (human: 74.3%, AI: 25.7%) in both splits.

Rather than applying undersampling or oversampling (e.g., SMOTE), class imbalance is addressed through cost-sensitive learning. For XGBoost and LightGBM, the *scale_pos_weight* parameter is set to the ratio of negative to positive instances:

$$w_+ = \frac{n_0}{n_1} \quad (5)$$

where $n_0$ and $n_1$ denote the counts of human and AI reviews in the training set, respectively. For Random Forest (RF) and Logistic Regression (LR), equivalent class weights are set via the *class_weight* parameter. This approach preserves the full training signal while penalizing misclassification of the minority class (AI-generated reviews) proportionally.

Grid search with 5-fold stratified cross-validation is performed for all four classifiers, optimizing for AUC-ROC. The search spaces are detailed in Table 4.

Table 4. Hyperparameter search spaces

| Classifier | Hyperparameter | Search range |
|---|---|---|
| *XGBoost* | n_estimators | {100, 200} |
| | max_depth | {3, 5, 7} |
| | learning_rate | {0.05, 0.1} |
| | subsample | {0.8, 1.0} |
| *RF* | n_estimators | {100, 200, 300} |
| | max_depth | {3, 5, 7, None} |
| | min_samples_split | {2, 5} |
| | min_samples_leaf | {1, 2} |
| *LightGBM* | n_estimators | {100, 200} |
| | max_depth | {3, 5, 7} |
| | learning_rate | {0.05, 0.1} |
| | num_leaves | {15, 31, 63} |
| *LR* | C (regularisation) | {0.01, 0.1, 1.0, 10.0, 100.0} |
| | l$_1$ ratio (ElasticNet) | {0.0, 0.5, 1.0} |

For the LR baseline, features are standardized using z-score normalization (StandardScaler) within a scikit-learn pipeline to prevent data leakage. The ElasticNet penalty (combining L$_1$ and L$_2$ regularization) is used with the SAGA solver and a maximum of 5,000 iterations to ensure convergence.

Model performance is assessed on the held-out test set using the following metrics. Accuracy, defined as the proportion of correctly classified reviews:

$$Acc = \frac{TP + TN}{TP + TN + FP + FN} \quad (6)$$

Precision, Recall and F1-score for each class:

$$Precision = \frac{TP}{TP + FP} \quad (7)$$

$$Recall = \frac{TP}{TP + FN} \quad (8)$$

$$F_1 = \frac{2 \cdot Precision \cdot Recall}{Precision + Recall} \quad (9)$$

Area Under the Receiver Operating Characteristic Curve (AUC-ROC) measures the classifier's ability to discriminate between classes across all probability thresholds. Additionally, confusion matrices are examined to assess false positive (human review misclassified as AI) and false negative (AI review misclassified as human) rates. These error types carry distinct practical implications: false positives unjustly flag legitimate human reviewers, while false negatives allow AI-generated reviews to pass undetected.

*3.5. SHAP-based explainability*

To ensure transparency and interpretability of the classification decisions, SHAP (SHapley Additive exPlanations) analysis is applied [39]. SHAP values are grounded in cooperative game theory and quantify the marginal contribution of each feature to an individual prediction. For a prediction $f(x_i)$, the SHAP value $\psi_j(x_i)$ for feature $j$ is defined as:

$$\psi_j(x_i) = \sum_{S \subseteq F \setminus \{j\}} \frac{|S|!(|F|-|S|-1)!}{|F|!} \cdot [f(S \cup \{j\}) - f(S)] \quad (10)$$

where $F=\{1, 2, ..., m\}$ is the set of all features, $S$ is a subset of features excluding $j$ and $f(S)$ denotes the model's expected output when only the features in $S$ are known. The SHAP values satisfy three desirable axioms: local accuracy (the sum of all SHAP values plus the base value equals the model output), missingness (features absent from the input receive zero attribution) and consistency (if a feature's contribution increases in a revised model, its SHAP value does not decrease).

For the tree-based classifiers (XGBoost, RF, LightGBM), the TreeExplainer algorithm is used for efficient exact computation of SHAP values in polynomial time, avoiding the exponential complexity of the general formula. Two levels of explainability are produced:

- *Global importance*-computed as the mean absolute SHAP value across all $N$ instances in the dataset, identifying which markers are most influential. Formally, the global importance of marker $j$ is $G_j=(1/N) \sum_i |\psi_j(x_i)|$;
- *Local explanations*-for individual reviews, showing how each marker pushed the prediction toward "Human" (negative SHAP value) or "AI-Generated" (positive SHAP value). The top-5 contributing features are reported with their signed SHAP values and direction indicators.

Both beeswarm and horizontal bar chart visualizations are further generated to indicate feature contributions to non-technical stakeholders such as journal editors and program committee chairs.

*3.6. Retrieval-augmented generation for evidence*

A Retrieval-Augmented Generation (RAG) module is developed to provide editors with evidence-based context alongside classifier predictions. The module implements a two-stage retrieval pipeline: embedding followed by approximate nearest-neighbor search. All $N=7,772$ reviews in the dataset are encoded into dense vector representations using the *all-MiniLM-L6-v2* sentence transformer [40], which maps variable-length text inputs to fixed-dimensional vectors $e_i \in \mathbb{R}^{nd}$, where $nd=384$. Encoding is performed in batches of 32 with $L_2$ normalization applied to all vectors:

$$\hat{e}_i = \frac{e_i}{\|e_i\|_2} \quad (11)$$

such that $\|\hat{e}_i\|_2=1$ for all $i$. The normalization enables the use of inner product as a proxy for cosine similarity. The normalized embeddings are indexed using FAISS (Facebook AI Similarity Search) [41] with an IndexFlatIP (flat inner-product) index. For normalized vectors, the inner product is equivalent to cosine similarity:

$$sim(\hat{e}_i, \hat{e}_j) = \hat{e}_i^T \cdot \hat{e}_j = cos(e_i, e_j) \quad (12)$$

The flat index performs exhaustive search over all $N$ vectors, guaranteeing exact nearest-neighbor results. While approximate indices (e.g., IVF, HNSW) offer faster search for larger corpora, exact search is chosen given the dataset size ($N=7,772$), where query latency is negligible.

For a query review $r_q$, the system computes its embedding $\hat{e}_q$ and retrieves the top-$K$=5 most similar reviews from the index, ranked by descending cosine similarity. To avoid trivial self-matches, any retrieved review whose text exactly matches the query is excluded, and $K$+1 candidates are initially retrieved to compensate. Each retrieved result includes the review text (truncated to 500 characters for display), the ground-truth label (Human or AI-Generated), the similarity score, the data source and the associated paper ID.

An aggregate summary is computed over the retrieved set, including: the total count of human and AI-generated matches, the average similarity score across the $K$ results and the label of the most similar review, serving as corroborative evidence: if the most similar reviews in the knowledge base are predominantly AI-generated, this corroborates the classifier's prediction and increases decision confidence.

*3.7. Automated editor report generation*

The final stage of the pipeline synthesizes all preceding outputs into a structured, actionable editor report. The report generation module follows a sequential processing flow: (1) extract marker scores using the LLM extractor (as in subsection 3.3), falling back to the rule-based extractor if the API is unavailable; (2) classify the review using the trained model; (3) compute SHAP explanations for the prediction; (4) retrieve similar reviews from the RAG index; and (5) combine all outputs into the final report.

The report includes the following components:

- *Classification result*-the predicted label (Human or AI-Generated), the AI probability $P(y=1|x)$, and a confidence level. Confidence is determined by the maximum class probability: High if $max(P_0, P_1)$>0.8, Medium if $max(P_0, P_1)$>0.6 and Low otherwise;
- *Marker scores with severity*-each of the eight marker scores is reported alongside a severity rating: High (score>0.7), Medium (0.4<score≤ 0.7) or Low (score≤ 0.4), providing editors with an intuitive sense of which aspects of the review triggered the detection;
- *SHAP explanation*-the top-5 features ranked by absolute SHAP value, each annotated with its signed contribution and direction ("toward AI" or "toward Human"). The SHAP base value is also reported, representing the model's average prediction across the training set;
- *RAG evidence*-the top-3 most similar reviews from the knowledge base, with their labels, similarity scores, text previews (first 200 characters) and source metadata. The aggregate summary (human vs. AI match counts, average similarity) provides a second line of evidence independent of the classifier;
- *Overall assessment*-a textual summary combining the AI probability and marker severity counts. Four assessment levels are defined: STRONG indicators (P>0.8 and ≥3 high-severity markers), MODERATE indicators (P>0.6 and ≥2 high markers), WEAK indicators (P>0.4) and "appears human-authored" (P≤0.4).

The assessment formula for the overall verdict integrates both the probabilistic classifier output and the marker-level evidence:

$$Assessment = g(P(y=1|x), |\{j : x_j > 0.7\}|) \quad (13)$$

where $g$ is a rule-based function that maps the AI probability and the count of high-severity markers to one of the four assessment levels, ensuring that the final recommendation is not driven solely by the classifier's point estimate but also reflects the breadth of evidence across multiple linguistic dimensions. Reports are exported in both JSON format (for programmatic consumption) and formatted plain text (for human reading).

Figure 1 presents the entire research methodology workflow.

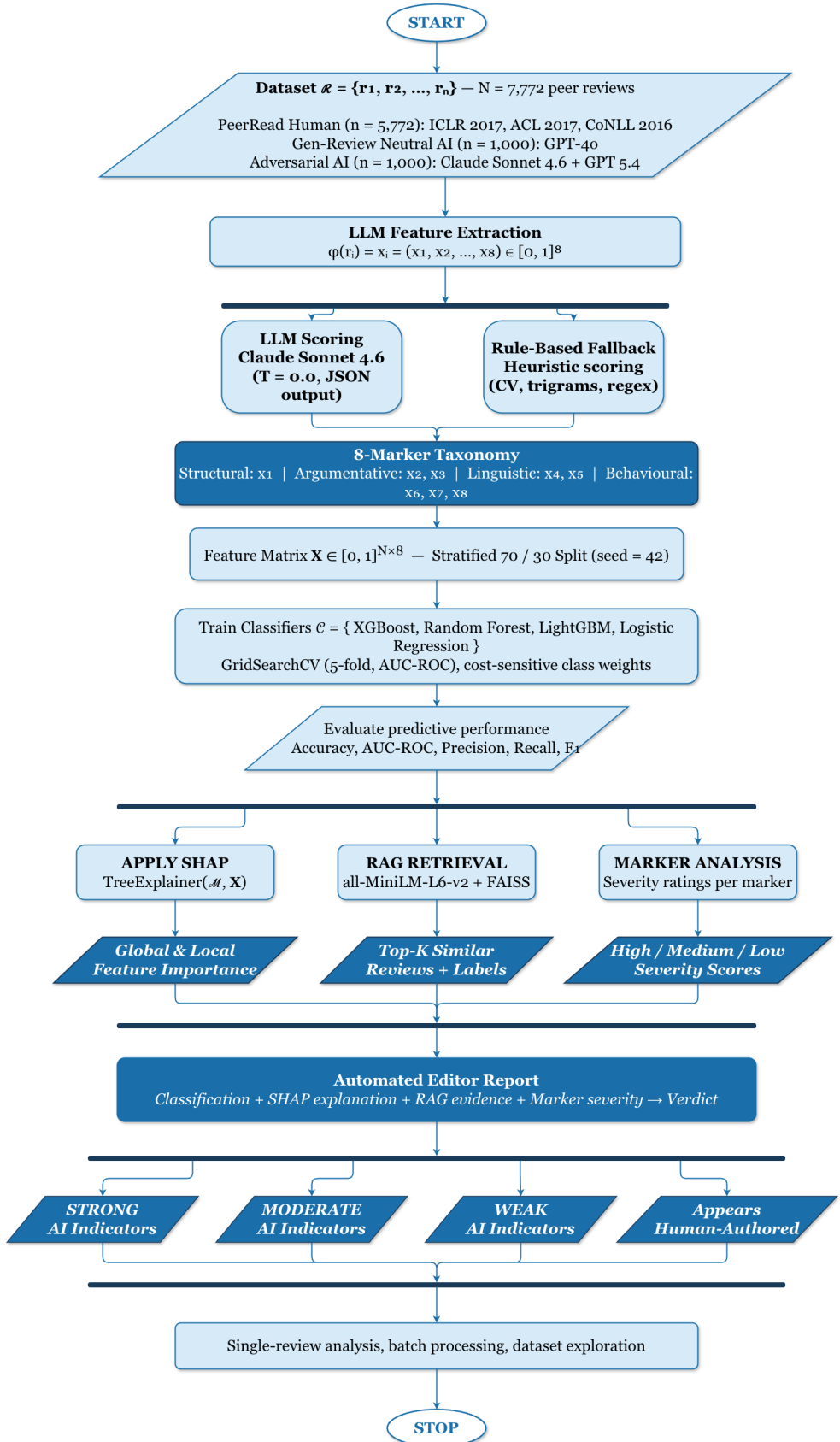

Figure 1. Research methodology workflow

## 4. Results

This section presents the experimental results of the proposed AI-generated peer review detection framework. We report the findings in five subsections: dataset overview, descriptive analysis of the marker distributions, classifier comparison on the full dataset, SHAP-based explainability analysis and RAG retrieval evaluation.

*4.1. Dataset overview*

Figure 2 provides a visual summary of the dataset composition: (a) shows the class distribution: 5,772 human-authored reviews and 2,000 AI-generated reviews, yielding a class ratio of approximately 74:26; (b) disaggregates the data by source, revealing that the human class is dominated by ICLR 2017 reviews (n=5,458), supplemented by smaller contributions from ACL 2017 (n=275) and CoNLL 2016 (n=39). The AI class comprises 1,000 neutral Gen-Review reviews and 1,000 adversarial reviews (500 from Claude Sonnet 4.6 and 500 from GPT 5.4); (c) shows the review length distributions: human reviews exhibit a right-skewed distribution peaking between 100 and 300 words, while AI-generated reviews concentrate in a narrower range around 500–800 words, reflecting the more consistent output length of LLM generation.

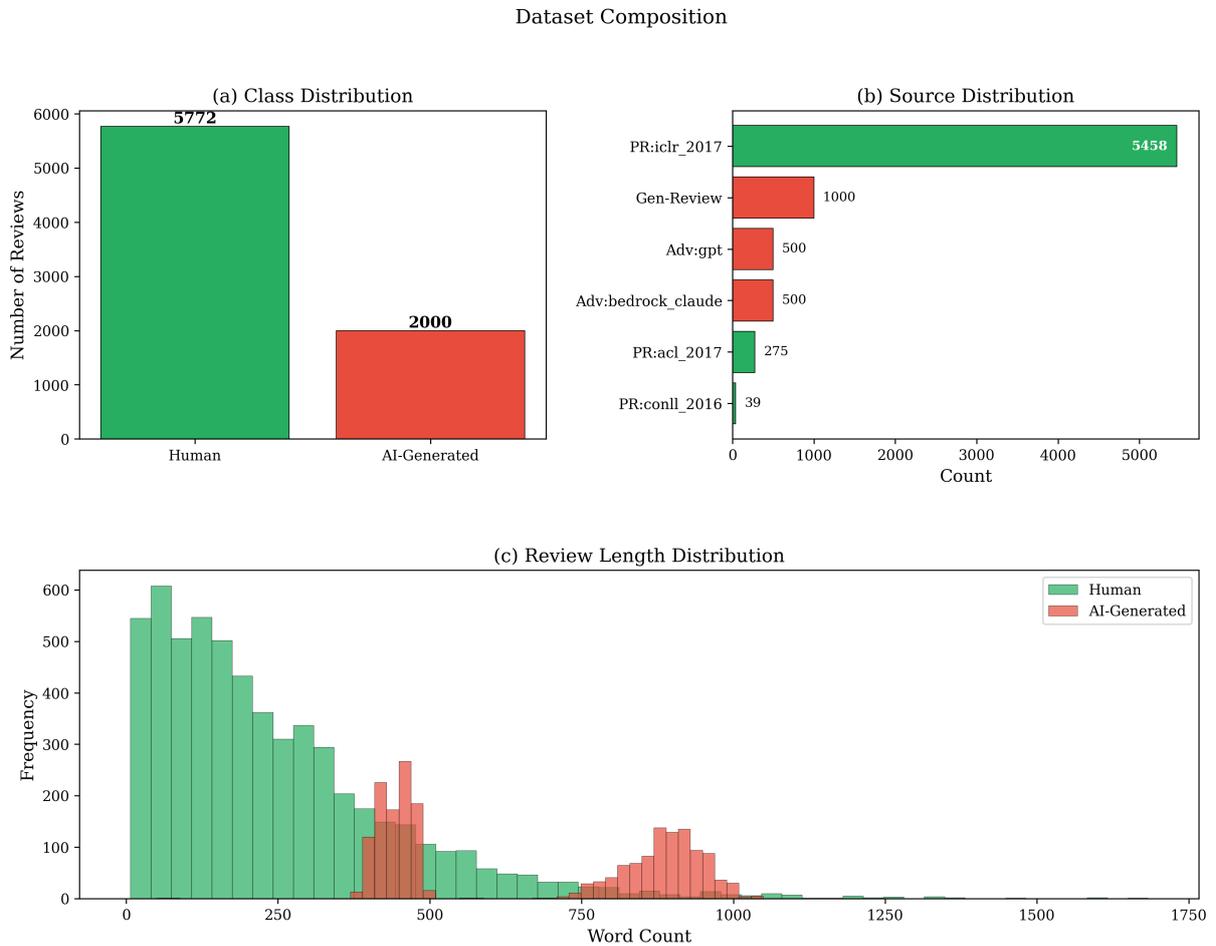

Figure 2. Dataset composition: (a) class distribution (Human vs. AI-Generated), (b) source distribution showing the contribution of each data subset, and (c) review length distributions by class

*4.2. Descriptive analysis of marker distributions*

Before training classifiers, we examine the distributional properties of the eight marker scores across the human and AI-generated review classes. Table 5 reports the mean and standard deviation of each marker, together with the mean difference Δ between AI-generated and human reviews.

Table 5. Marker descriptive statistics by class (mean ± standard deviation)

| Marker | Human (n=5,772) | AI (n=2,000) | Δ (AI−Human) |
|---|---|---|---|
| *Standardized structure ($x_1$)* | 0.205±0.227 | 0.457±0.393 | +0.252 |
| *Predictable criticism ($x_2$)* | 0.213±0.146 | 0.427±0.249 | +0.215 |
| *Excessive balance ($x_3$)* | 0.268±0.178 | 0.512±0.243 | +0.244 |
| *Linguistic homogeneity ($x_4$)* | 0.524±0.128 | 0.554±0.258 | +0.030 |
| *Generic domain language ($x_5$)* | 0.284±0.148 | 0.473±0.323 | +0.189 |
| *Conceptual Feedback ($x_6$)* | 0.502±0.208 | 0.581±0.266 | +0.079 |
| *Absence Personal Signals ($x_7$)* | 0.405±0.265 | 0.469±0.425 | +0.065 |
| *Repetition Patterns ($x_8$)* | 0.167±0.112 | 0.398±0.300 | +0.231 |

First, all eight markers show higher mean values for AI-generated reviews compared to human reviews, confirming the directional hypotheses underlying the taxonomy. Second, the magnitude of separation varies considerably across markers. *Standardized structure* (Δ=+0.252), *Excessive balance* (Δ=+0.244), and *Repetition patterns* (Δ=+0.231) exhibit the largest mean differences, suggesting these are the most discriminative markers at the population level.

Figure 3 presents the per-marker score distributions as overlaid histograms for human (green) and AI-generated (red) reviews. The distributions reveal that markers such as *Standardized structure* and *Repetition patterns* show clearly separated modes between the two classes, while *Linguistic homogeneity* and *Conceptual feedback* exhibit substantial overlap, indicating that these markers alone provide weaker class separation. Notably, several AI-marker distributions are bimodal, reflecting the mixture of neutral and adversarial AI reviews.

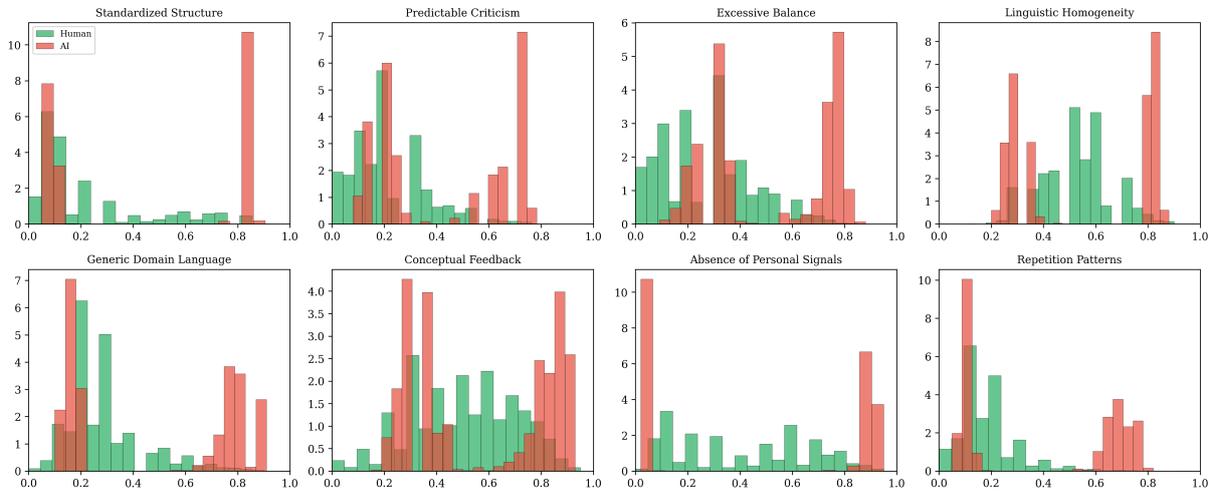

Figure 3. Per-marker score distributions for human (green) and AI-generated (red) reviews, shown as overlaid histograms for each of the eight markers

Figure 4 provides a complementary view through box plots, which summarize the median, interquartile range and outliers for each marker by class. The box plots confirm the pattern observed in the histograms: markers with large mean differences (*Standardized structure*, *Repetition patterns*) show non-overlapping interquartile ranges, while markers with smaller differences (*Linguistic homogeneity*, *Conceptual feedback*) exhibit overlapping boxes but still differ in their median values.

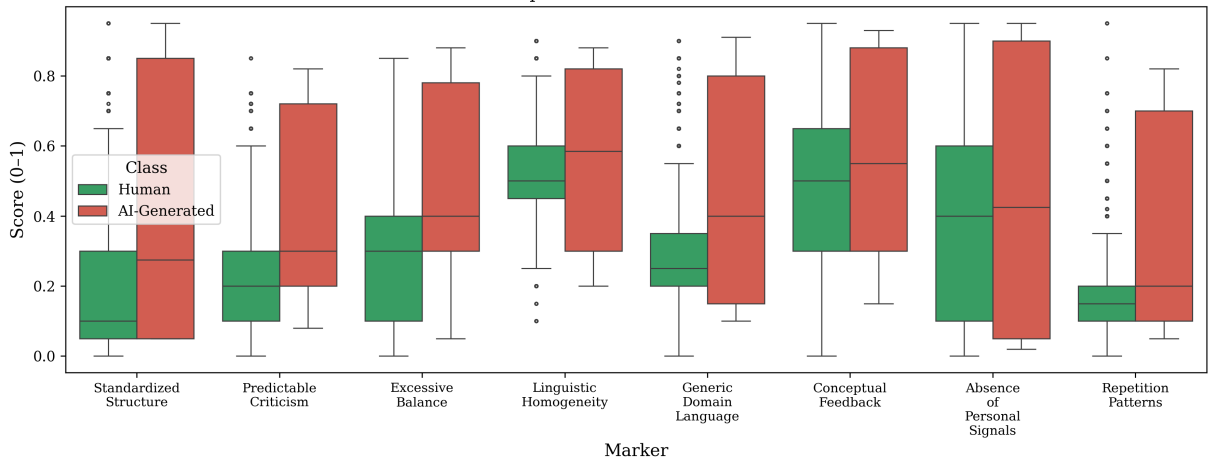

Figure 4. Box plot comparison of the eight marker scores by class (Human vs. AI-Generated). Boxes indicate the interquartile range (IQR); whiskers extend to 1.5×IQR; individual points indicate outliers

Notably, *Linguistic homogeneity* (Δ=+0.030) and *Absence of personal signals* (Δ=+0.065) show relatively small differences in their overall means. However, these aggregate statistics mask an important compositional effect: the adversarial AI reviews are specifically designed to minimize these markers, while the neutral reviews extracted through the Gen-Review framework exhibit extremely high scores (e.g., *Absence of personal signals*: 0.894 for neutral AI vs. 0.048 for adversarial Claude). The bimodal distribution within the AI class reduces the aggregate mean difference but does not diminish the marker's discriminative power for non-adversarial detection. Table 6 disaggregates the AI-generated reviews by source to illustrate the compositional effect.

Table 6. Marker means by AI data source

| Marker | Gen-Review (neutral, n=1,000) | Adversarial Claude 4.6 (n=500) | Adversarial GPT 5.4 (n=500) |
|---|---|---|---|
| *Standardized structure* | 0.849 | 0.077 | 0.053 |
| *Predictable criticism* | 0.665 | 0.201 | 0.177 |
| *Excessive balance* | 0.748 | 0.262 | 0.292 |
| *Linguistic homogeneity* | 0.810 | 0.298 | 0.299 |
| *Generic domain language* | 0.793 | 0.164 | 0.143 |
| *Conceptual feedback* | 0.840 | 0.330 | 0.315 |
| *Absence personal signals* | 0.894 | 0.048 | 0.042 |
| *Repetition patterns* | 0.696 | 0.105 | 0.096 |

The contrast is evident. Neutral Gen-Review reviews score above 0.65 on all eight markers, with *Absence of personal signals* (0.894) and *Standardized structure* (0.849) reaching near-ceiling values. In contrast, both adversarial sources produce marker scores that are comparable to, or even lower than, human reviews on several markers (e.g., *Standardized structure*: 0.077 for adversarial Claude vs. 0.205 for human), confirming that the adversarial prompts successfully induced human-mimicking behavior, creating a genuinely challenging detection scenario. The two adversarial sources (Claude Sonnet 4.6 and GPT 5.4) produce remarkably similar marker profiles, suggesting that the evasion strategies generalize across LLM architectures.

Figure 5 presents the inter-marker correlation matrix. The highest pairwise correlations are observed between *Repetition patterns* and *Standardized structure* (r=0.83) and between *Predictable criticism* and *Generic domain language* (r=0.83), indicating that reviews with formulaic structure also tend to exhibit repetitive phrasing and that generic domain vocabulary co-occurs with *Predictable criticism* patterns. *Generic domain language* and *Conceptual feedback* also show strong correlation (r=0.81), suggesting that reviews relying on domain-generic vocabulary tend to provide more superficial, template-driven feedback. *Linguistic homogeneity* and *Absence of personal signals* (r=0.78), as well as *Excessive balance* and *Generic domain language* (r=0.78), form another cluster of moderately correlated features. The lowest pairwise correlations involve *Excessive balance* and *Absence of personal signals* (r=0.41) and *Conceptual feedback* and *Standardized structure* (r=0.41), confirming that these markers capture complementary signal dimensions. The generally moderate-to-strong positive correlations indicate that AI-generated text tends to trigger multiple markers simultaneously, while the absence of near-unity correlations confirms that no two markers are redundant.

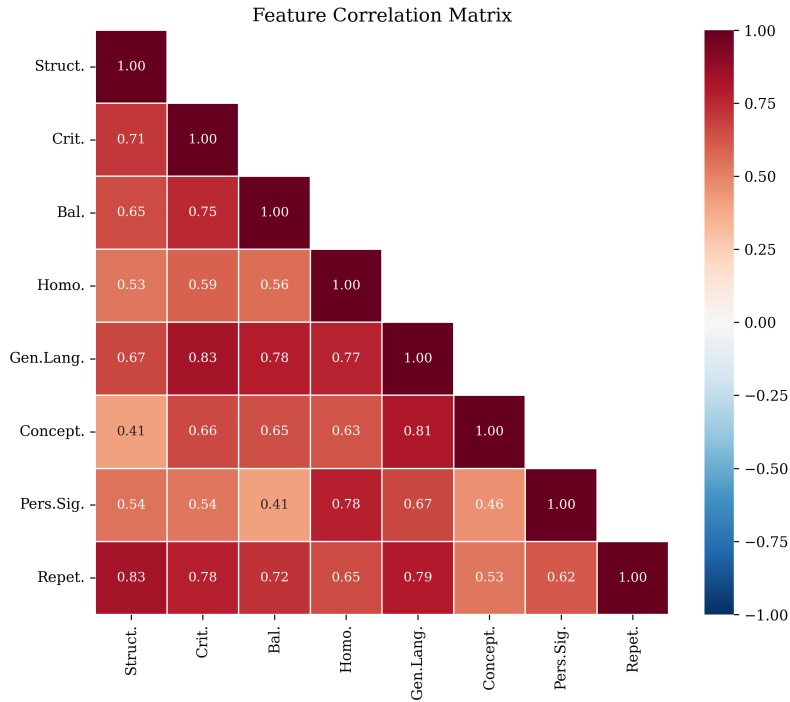

Figure 5. Pearson correlation matrix of the eight marker scores computed over the full dataset

*4.3. Classifier comparison*

Table 7 presents the performance of the four classifiers on the held-out test set (*n*=2,332). The three ensemble methods, XGBoost, RF and LightGBM, achieve near-identical performance, while the LR baseline performs substantially lower.

Table 7. Classifier performance on the held-out test set (*n*=2,332)

| Classifier | Accuracy | AUC-ROC | Precision | Recall | F1-score |
|---|---|---|---|---|---|
| *XGBoost* | 0.9961 | 0.9996 | 0.9933 | 0.9917 | 0.9925 |
| *RF* | 0.9961 | 0.9997 | 0.9950 | 0.9900 | 0.9925 |
| *LightGBM* | 0.9961 | 0.9997 | 0.9950 | 0.9900 | 0.9925 |
| *LR* | 0.8997 | 0.9584 | 0.7322 | 0.9617 | 0.8314 |

The three tree-based ensemble classifiers achieve an identical accuracy of 99.61% and AUC-ROC values exceeding 0.999, demonstrating that the eight-marker feature representation provides sufficient discriminative power for near-perfect binary classification. The marginal differences between XGBoost, RF and LightGBM are not statistically meaningful at this performance level.

Figure 6 provides a visual comparison of three performance metrics: accuracy, AUC-ROC and F1-score, across classifiers through grouped bar charts: (a) and (b) confirm the near-ceiling performance of the tree-based ensembles (all above 0.99), while the LR baseline shows a visible gap (0.8997 accuracy and 0.9584 AUC-ROC); (c) highlights the F1-score disparity most starkly: all three ensemble methods achieve 0.9925, whereas LR falls to 0.8314, underscoring the precision deficit of the linear model.

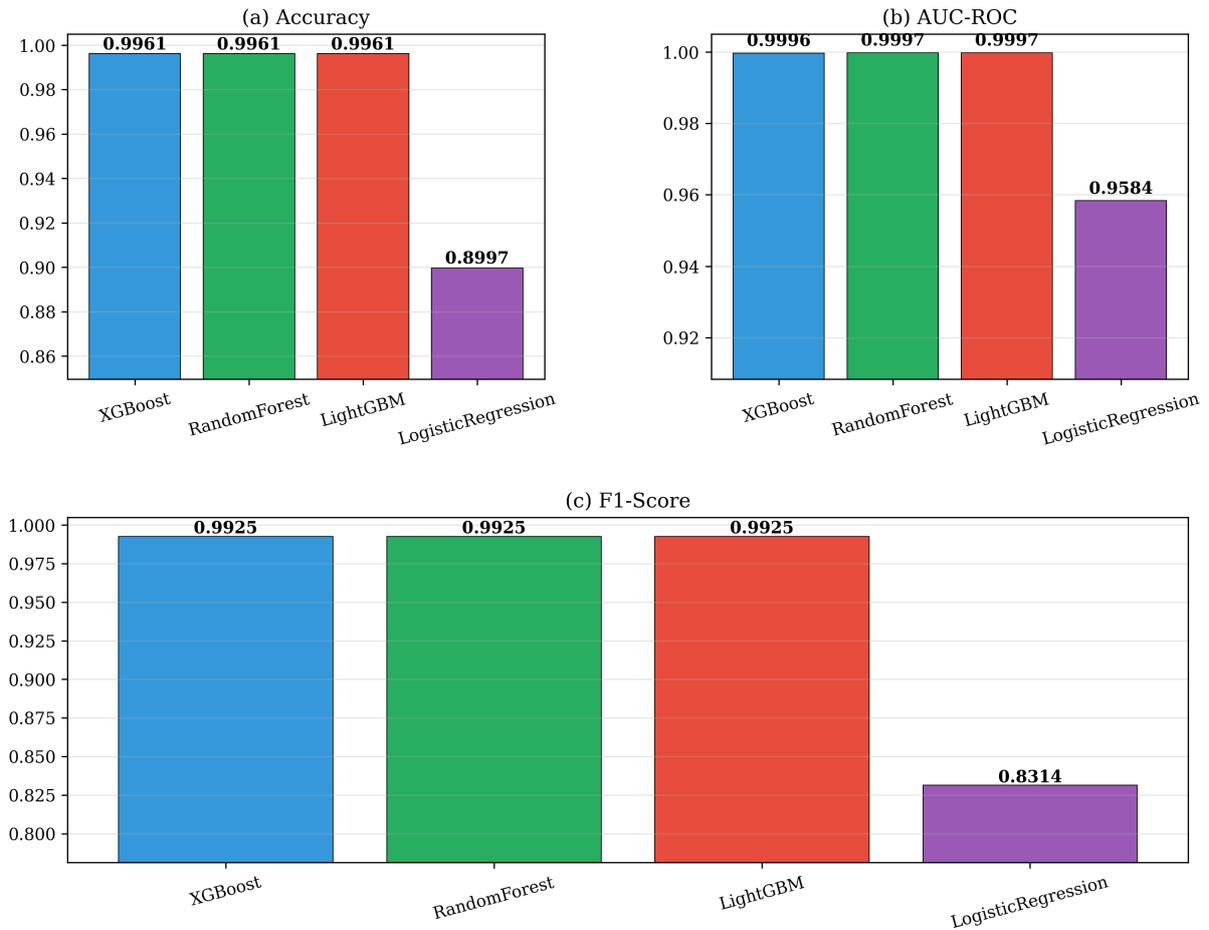

Figure 6. Classifier performance comparison: (a) Accuracy, (b) AUC-ROC and (c) F1-score for all four classifiers.

LR, serving as a linear baseline, achieves a considerably lower accuracy of 89.97% and AUC-ROC of 0.958. While the recall is high (0.962), the precision drops to 0.732, resulting in an F1-score of 0.831, suggesting that the decision boundary between human and AI reviews in the eight-dimensional marker space is non-linear, which the ensemble methods capture effectively through their tree-based partitioning.

Figure 7 presents the ROC curves for all four classifiers. The three ensemble methods, XGBoost (AUC=0.9996), RF (AUC=0.9997) and LightGBM (AUC=0.9997), produce near-rectangular curves that hug the top-left corner, indicating near-perfect discrimination across all classification thresholds. The three curves are virtually indistinguishable, reflecting the equivalent performance of the ensemble methods. LR (AUC=0.9584) exhibits a visibly different trajectory: while it achieves high true positive rates, it does so at the cost of substantially higher false positive rates, with the curve deviating noticeably from the top-left corner. The dashed diagonal represents the random baseline (AUC=0.5), against which all classifiers show a clear separation.

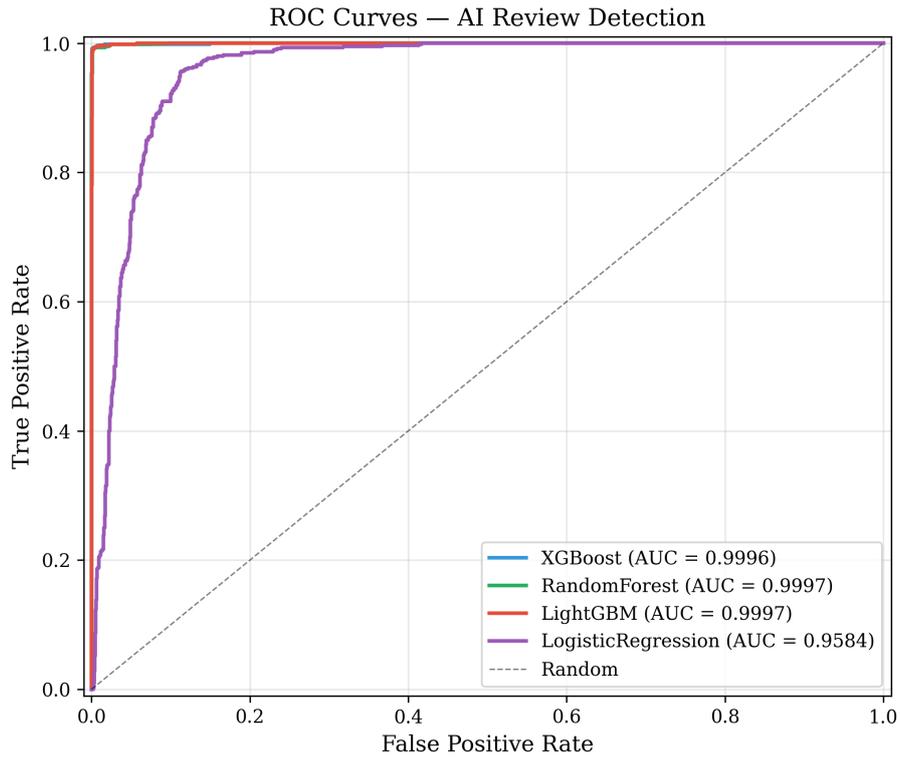

Figure 7. ROC curves for all four classifiers, with AUC scores shown in the legend

Table 8 and Figure 8 report the confusion matrices for all four classifiers. The ensemble methods produce remarkably few errors: XGBoost misclassifies only 4 human reviews as AI (false positives) and 5 AI reviews as human (false negatives), out of 2,332 test instances. RF and LightGBM each produce 3 false positives and 6 false negatives.

Table 8. Confusion matrices on the held-out test set

| Classifier | TP | FP | FN | TN | FPR | FNR |
|---|---|---|---|---|---|---|
| *XGBoost* | 595 | 4 | 5 | 1,728 | 0.23% | 0.83% |
| *RF* | 594 | 3 | 6 | 1,729 | 0.17% | 1.00% |
| *LightGBM* | 594 | 3 | 6 | 1,729 | 0.17% | 1.00% |
| *LR* | 577 | 211 | 23 | 1,521 | 12.18% | 3.83% |

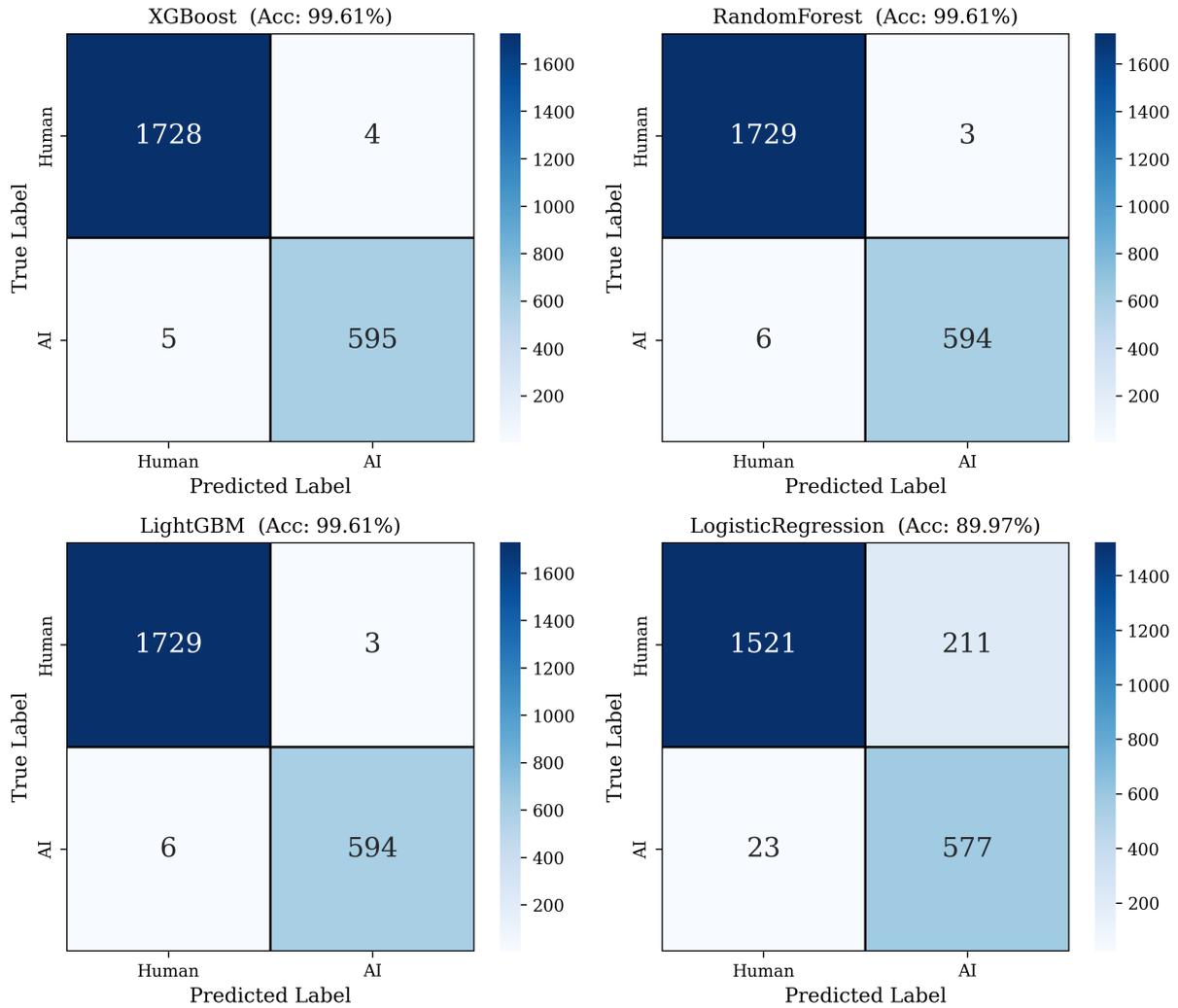

Figure 8. Confusion matrices for all four classifiers on the held-out test set

The false positive rate (FPR) of the ensemble methods is below 0.23%, meaning that fewer than 1 in 400 legitimate human reviews would be incorrectly flagged as AI-generated. This is important as a high false positive rate would undermine trust in the detection system by wrongly accusing human reviewers. The false negative rate (FNR) of approximately 0.8% indicates that a small number of AI-generated reviews may evade detection, which is expected given the adversarial subset.

In contrast, LR produces 211 false positives (FPR=12.18%), which would be unacceptable in a production deployment where over one in ten legitimate reviews would be flagged, underscoring the need for non-linear classifiers in this detection task.

### 4.4. Feature importance and SHAP analysis

To understand which markers drive classifier decisions, we examine two complementary perspectives: *(i)* native model feature importance and *(ii)* SHAP-based explainability analysis. Table 9 reports the native feature importance for XGBoost, which reflects the aggregate gain contributed by each feature across all trees.

Table 9. Native feature importance, XGBoost (gain-based)

| Rank | Marker | Importance | Relative contribution |
|---|---|---|---|
| 1 | Absence of personal signals ($x_7$) | 0.5782 | 57.8% |
| 2 | Repetition patterns ($x_8$) | 0.2939 | 29.4% |
| 3 | Conceptual feedback ($x_6$) | 0.0275 | 2.7% |
| 4 | Standardized structure ($x_1$) | 0.0266 | 2.7% |
| 5 | Linguistic homogeneity ($x_4$) | 0.0238 | 2.4% |
| 6 | Predictable criticism ($x_2$) | 0.0176 | 1.8% |
| 7 | Generic domain language ($x_5$) | 0.0162 | 1.6% |
| 8 | Excessive balance ($x_3$) | 0.0161 | 1.6% |

*Absence of personal signals* emerges as the most influential marker in the native XGBoost importance, contributing 57.8% of the total gain, followed by *Repetition patterns* (29.4%), both account for over 87% of the model's total feature importance, indicating that the XGBoost ensemble heavily relies on these features to partition the decision space.

The native importance ranking, however, captures only aggregate gain, a measure of how frequently and effectively a feature is used for splitting across all trees. It does not account for feature interactions or the direction of effect. To address these limitations, we complement the native importance analysis with SHAP (SHapley Additive exPlanations), which provides instance-level, theoretically grounded attribution scores.

Figure 9 presents the global SHAP feature importance (mean |SHAP value|) side-by-side for all three tree-based classifiers, colour-coded by marker category (Structural, Argumentative, Linguistic, Behavioural). Across all three models, SHAP assigns the highest importance to *Absence of Personal Signals*, with mean |SHAP| values of 2.652 (XGBoost), 0.121 (RandomForest) and 3.508 (LightGBM). The native gain-based importance rankings are broadly consistent, also placing *Absence of Personal Signals* first. The alignment arises because SHAP measures the average marginal contribution of each feature to individual predictions, accounting for feature correlations and interactions, while native importance measures aggregate split gain. The high native importance of *Absence of Personal Signals* reflects its frequent use as a splitting variable, but its redundancy with other markers (r=0.78 with *Linguistic Homogeneity*) means that its marginal SHAP contribution is partially absorbed by the correlated features.

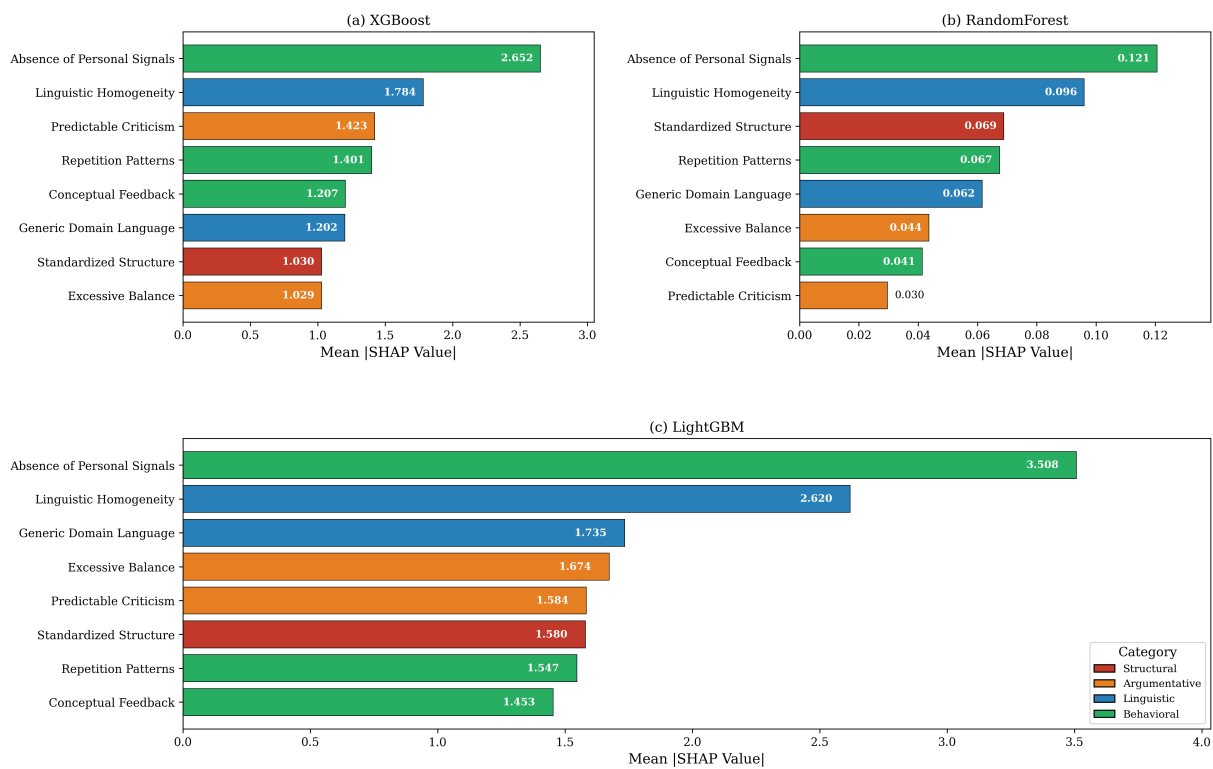

Figure 9. Global SHAP feature importance (mean |SHAP value|) for all three tree-based classifiers, color-coded by marker category

*Linguistic Homogeneity* consistently ranks second across all three tree-based classifiers in the SHAP analysis, with mean |SHAP| values of 1.784 (XGBoost), 0.096 (RandomForest) and 2.620 (LightGBM). This positions it as the second most discriminative marker after *Absence of Personal Signals*, reinforcing its role as a key linguistic signal for distinguishing AI-generated from human-written reviews. The absolute SHAP magnitudes differ substantially across classifiers: LightGBM produces the largest values (up to 3.508), reflecting its deeper-boosted tree structure, while RF operates on a much smaller scale (maximum 0.121) due to its probability-based output aggregation. Despite these scale differences, the relative feature rankings are broadly consistent, confirming that the same markers drive predictions regardless of the specific ensemble method.

Figures 10–12 present the SHAP beeswarm plots for XGBoost, RF and LightGBM, respectively. Each dot represents a single review; the horizontal position indicates the SHAP value (impact on the model's log-odds output) and the color encodes the marker score (red=high, blue=low). These plots reveal the direction of each marker effect: high marker scores (red dots on the right) consistently push the model's prediction toward the AI class, while low marker scores (blue dots on the left) push toward the human class.

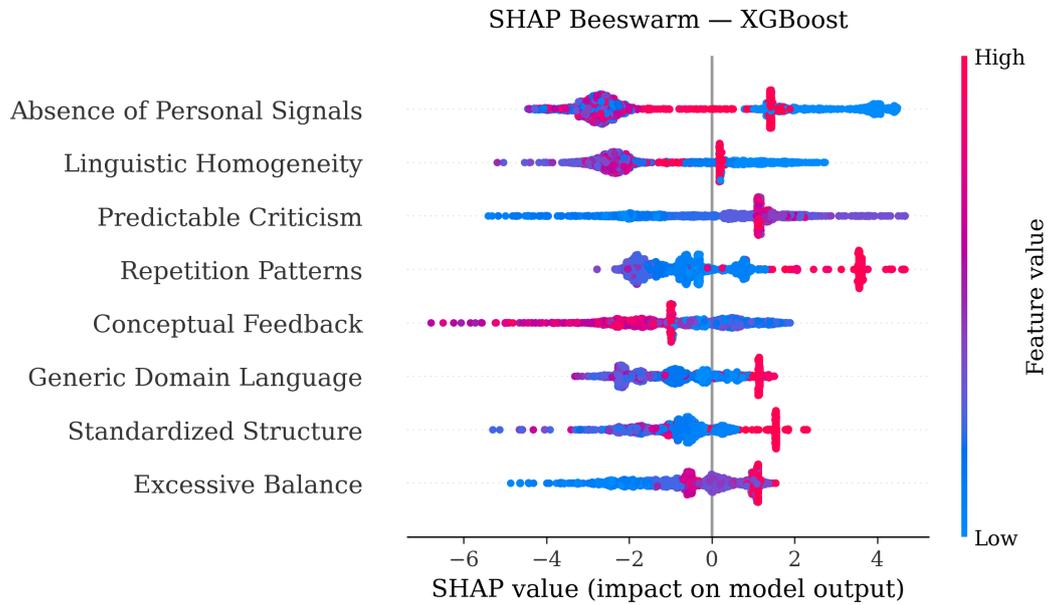

Figure 10. SHAP beeswarm plot, XGBoost. Each dot represents one review; horizontal position indicates the SHAP value and color encodes the feature value (red=high, blue=low)

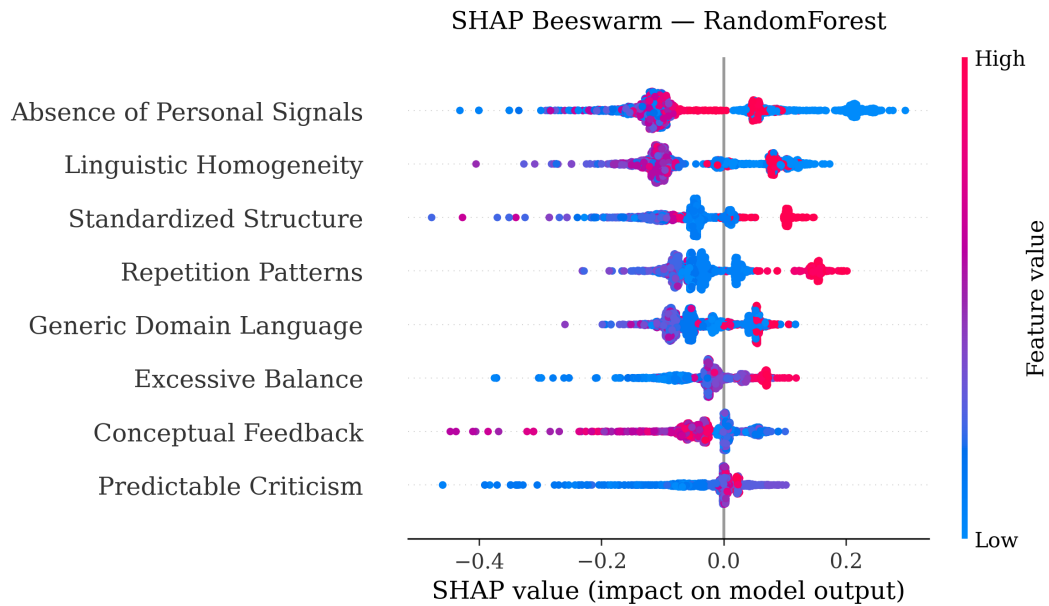

Figure 11. SHAP beeswarm plot, RF

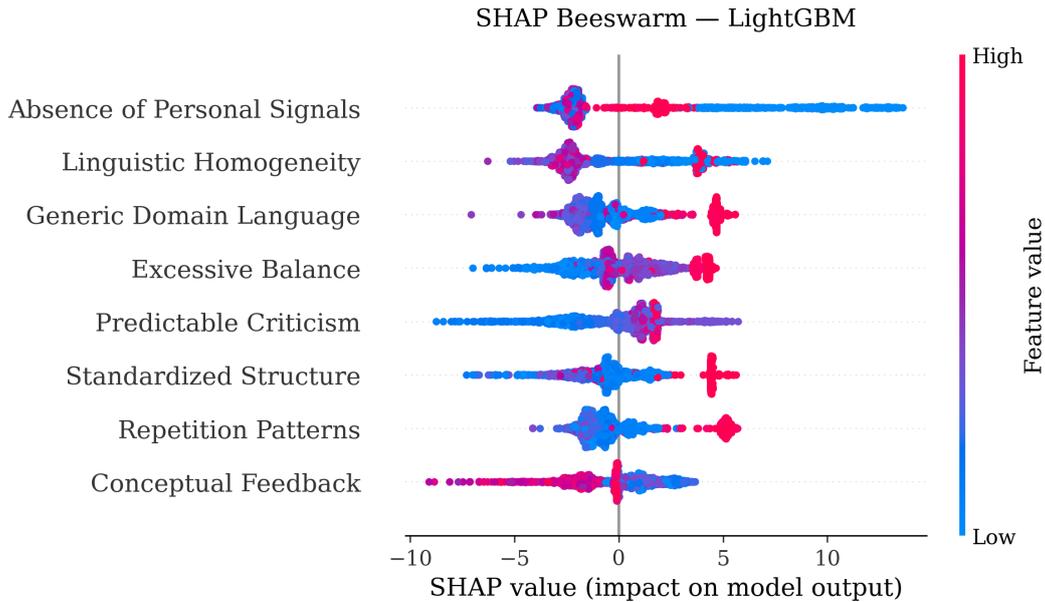

Figure 12. SHAP beeswarm plot, LightGBM

A consistent pattern across all three beeswarm plots is the strong bidirectional effect of *Absence of Personal Signals*: reviews with high values on this marker (red) receive large positive SHAP values, decisively pushing the prediction toward AI, while reviews with low values (blue) receive large negative SHAP values, anchoring the prediction to the human class. The XGBoost beeswarm (Figure 10) shows *Absence of Personal Signals* SHAP values spanning from approximately −7 to +5, by far the widest spread of any marker. *Linguistic Homogeneity* and *Predictable Criticism* exhibit similar directional patterns, though with more moderate magnitudes. In contrast, the bottom-ranked markers (*Standardized Structure*, *Excessive Balance*) show tightly clustered dots near zero, confirming their negligible marginal SHAP contribution. A notable difference across models emerges in the RF beeswarm (Figure 11). While *Absence of Personal Signals* and *Linguistic Homogeneity* retain the top two positions, *Standardized Structure* rises to third place (above *Repetition Patterns*) and *Conceptual Feedback* drops to seventh position, suggesting that the bagged ensemble captures complementary feature interactions that the boosted methods subsume into the top markers.

The LightGBM beeswarm (Figure 12) exhibits the largest SHAP magnitudes overall, with *Absence of Personal Signals* values extending from approximately −10 to +15, reflecting the model's deeper tree structure and more extreme log-odds.

In a cross-classifier comparison, Figure 13 presents the normalized feature importance comparison across all four classifiers (including LR). The feature achieving the maximum normalized importance (1.0) differs by model: *Absence of Personal Signals* dominates in XGBoost and RandomForest (both ≈1.0), while *Conceptual Feedback* reaches 1.0 in LightGBM and *Linguistic Homogeneity* ranks highest in LR (≈0.96). A key divergence appears in the secondary markers: tree-based methods show steep drop-offs after the top feature, whereas LR distributes importance more uniformly across all eight markers. In particular, *Linguistic Homogeneity* receives a normalized score of ≈0.96 in LR but varies widely across tree-based models (0.03 for XGBoost vs. 0.67 for RF and 0.93 for LightGBM), reflecting the linear model's tendency to assign stable weights to individually predictive features. Conversely, *Absence of Personal Signals* dominates the tree-based boosting and bagging ensembles but receives only ≈0.42 in LR, suggesting that its discriminative power is best captured through nonlinear interactions. LightGBM stands out for distributing high importance broadly, with five features exceeding 0.55, whereas XGBoost concentrates importance heavily on *Absence of Personal Signals* and penalizes other features more aggressively.

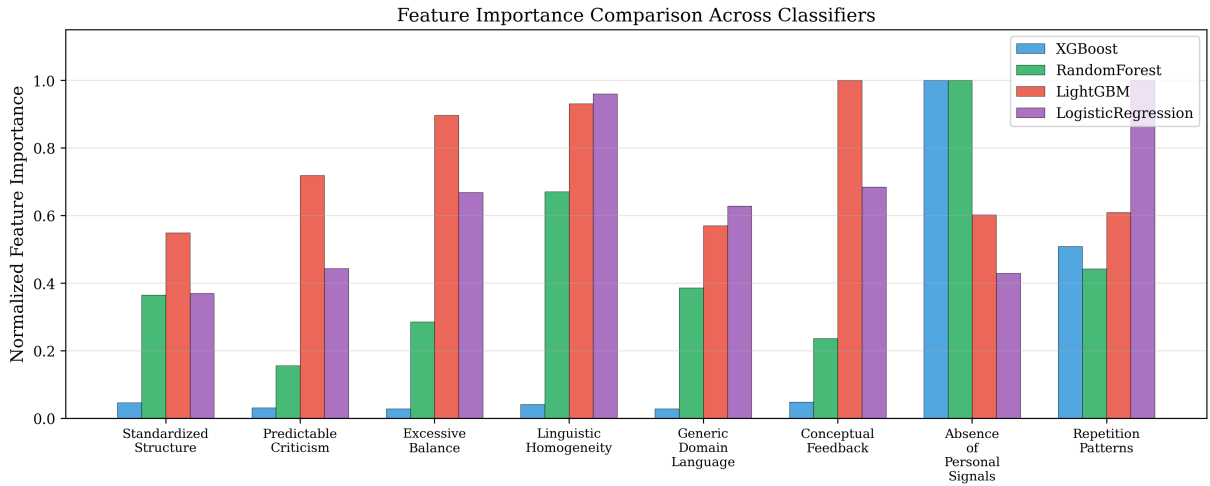

Figure 13. Normalized feature importance comparison across all four classifiers. Values are scaled relative to the maximum importance within each model to enable cross-model comparison

*4.5. RAG retrieval evaluation*

The final stage of the framework employs a Retrieval-Augmented Generation (RAG) application to provide evidence-based support for the classifier's predictions. For each query review, the RAG module retrieves the most similar reviews from the knowledge base using FAISS with sentence-transformer embeddings (*all-MiniLM-L6-v2*). The retrieved reviews serve as contextual evidence: if a review classified as AI-generated is most similar to other AI-generated reviews in the knowledge base, this corroborates the classifier's decision and provides interpretable examples for the editor.

To evaluate the RAG component, we sampled 100 human and 100 AI-generated reviews as queries and retrieved the top-5 most similar reviews from the full knowledge base (excluding the query itself). Table 10 reports the retrieval metrics.

Table 10. RAG retrieval evaluation metrics

| Metric | Human queries (n=100) | AI queries (n=100) |
|---|---|---|
| *Top-1 accuracy* | 91% | 90% |
| *Avg same-class in top-5* | 3.2 | 4.0 |
| *Avg cross-class in top-5* | 0.3 | 1.0 |
| *Mean top-5 similarity* | 0.661 | 0.629 |
| *Mean top-1 similarity* | 0.675 | 0.666 |

The top-1 retrieval accuracy, defined as the proportion of queries whose nearest neighbor shares the same class label, reaches 91% for human queries and 90% for AI-generated queries, yielding an overall accuracy of 90.5% (Figure 14). It indicates that the semantic embedding space naturally clusters reviews by origin, even though the embeddings are computed from raw review text without access to the marker scores or classifier labels.

The top-5 composition analysis further confirms this clustering effect. For human queries, an average of 3.2 out of the top-5 retrieved reviews are human-authored, with only 0.3 AI-generated reviews appearing in the neighborhood. For AI queries, the pattern mirrors: an average of 4.0 of the top-5 neighbors are AI-generated, with 1.0 human reviews, providing a secondary signal that reinforces the classifier's binary prediction.

The mean cosine similarity scores are comparable across query types: 0.661 for human queries and 0.629 for AI-generated queries, suggesting that both classes exhibit similar levels of intra-class semantic cohesion when measured through general-purpose sentence embeddings. The slightly higher similarity for human queries likely reflects the larger pool of human reviews in the knowledge base (5,772 vs. 2,000), which increases the likelihood of finding close semantic matches.

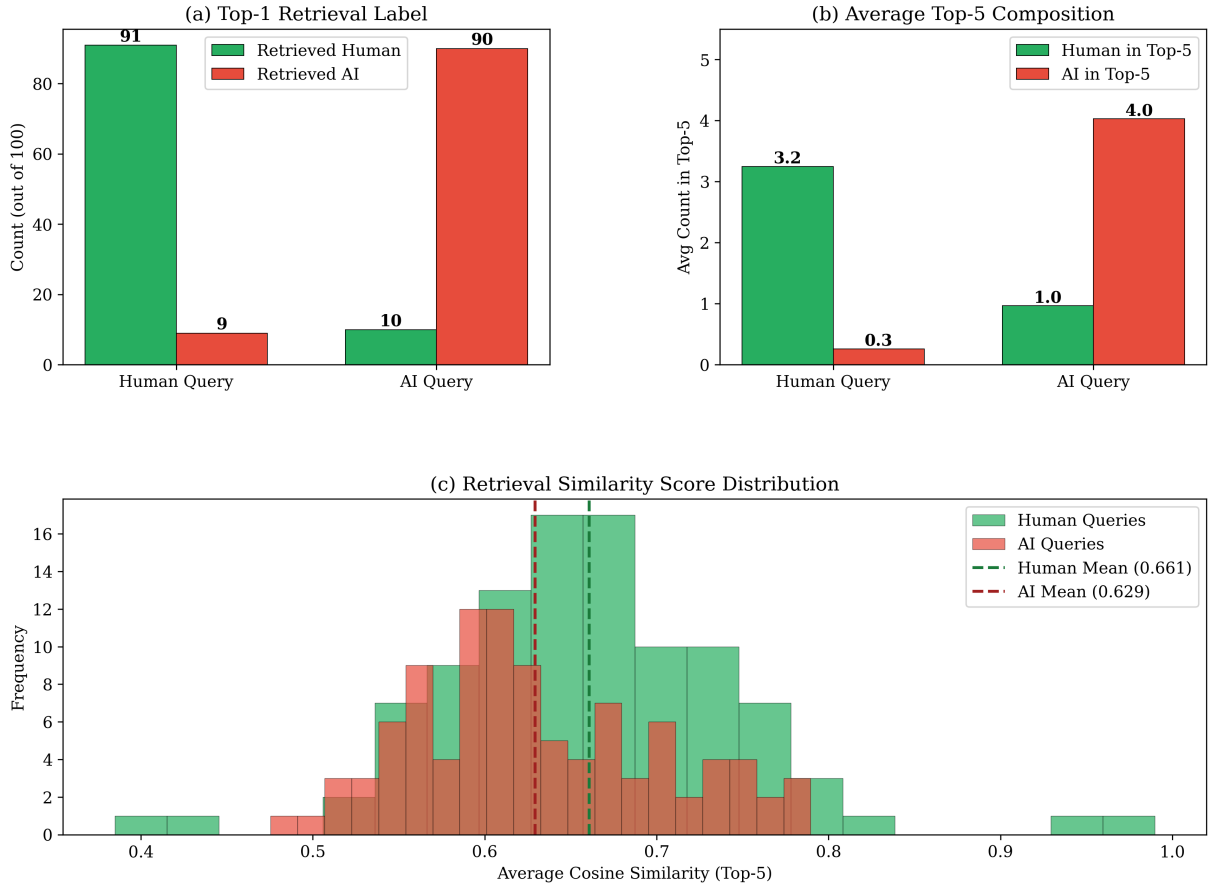

Figure 14. RAG retrieval evaluation: (a) top-1 retrieval label accuracy for human and AI queries, (b) average top-5 neighborhood composition by class, and (c) distribution of average cosine similarity scores across queries

The RAG module does not contribute to the binary classification decision itself; rather, it serves an interpretability and evidence function. When the classifier flags a review as potentially AI-generated, the retrieved similar reviews provide concrete examples that an editor can examine to understand why the review was flagged. The high top-1 accuracy demonstrates that the retrieved evidence is typically class-consistent, meaning the examples shown to the editor are representative of the predicted class. In the small number of cases where the nearest neighbor is from the opposite class, this may itself be informative, suggesting a boundary case that warrants closer editorial scrutiny.

## 5. Conclusions

The experimental results demonstrate that the proposed AI-generated peer review detection framework is effective and deployable. The eight-marker taxonomy exhibits strong discriminative power, with all markers consistently yielding higher values for AI-generated reviews than for human-authored ones, thereby validating the underlying theoretical assumptions. Among these, *standardized structure, excessive balance* and *repetition patterns* emerge as the most discriminative at the aggregate level, while other markers such as *linguistic homogeneity* and *conceptual feedback* contribute complementary information despite partial overlap. Importantly, the inclusion of adversarially generated reviews confirms the framework's robustness under realistic conditions, as these samples successfully mimic human-like patterns and reduce marker separability, creating a genuinely challenging detection scenario.

The classification results further reinforce the effectiveness of the approach. Tree-based ensemble models (XGBoost, RF and LightGBM) achieve near-perfect performance, with accuracy exceeding 99.6% and AUC-ROC values approaching 1.0, while maintaining extremely low false positive and false negative rates. In contrast, the significantly weaker performance of LR highlights the inherently non-linear nature of the decision boundary in the marker space, emphasizing the necessity of ensemble methods. Explainability analyses using both native feature importance and SHAP confirm that the *absence of personal signals* and *repetition patterns* are the dominant drivers of model predictions, with consistent directional effects across all classifiers. These findings also reveal that AI-generated reviews are characterized by a combination of structural regularity, reduced personalization and repetitive phrasing, rather than any single isolated feature.

Moreover, the RAG-based retrieval component demonstrates strong semantic clustering, achieving a top-1 accuracy of approximately 90.5% and consistently retrieving same-class neighbors for both human and AI queries. While not directly contributing to classification, this component enhances the interpretability and trustworthiness of the system by providing concrete, evidence-based examples that support model decisions. Overall, the results indicate that the proposed framework achieves state-of-the-art detection performance and addresses practical requirements such as robustness to adversarial manipulation, model interpretability and human-centered validation, making it well-suited for deployment in real peer review monitoring and editorial decision-support systems.

The increasing use of AI-assisted or AI-curated peer reviews presents a challenge that is difficult to regulate or prevent, as such usage is often indistinguishable from human-written content and can be easily disguised through editing or paraphrasing. Consequently, efforts to detect whether a review was generated or assisted by AI may be inherently limited and ultimately ineffective at scale. A more practical and future approach is therefore to shift the focus from authorship detection to evaluation of review quality and relevance, assessing whether the feedback meaningfully engages with the manuscript's content, methodology and contributions. By prioritizing criteria such as specificity, technical depth, consistency and actionable insight, the peer-review process can be safeguarded against low-quality or generic assessments, regardless of whether they originate from humans or AI, thus promoting a more outcome-oriented and resilient evaluation framework.

**Declarations**
Funding-This work was supported by a grant of the Ministry of Research, Innovation and Digitization, CNCS/CCCDI-UEFISCDI, project number COFUND-CETP-SMART-LEM-1, within PNCDI IV.
Acknowledgement-This work was supported by a grant of the Ministry of Research, Innovation and Digitization, CNCS/CCCDI-UEFISCDI, project number COFUND-CETP-SMART-LEM-1, within PNCDI IV.
Conflicts of interest/Competing interests-The authors declare that there is no conflict of interest.
Ethics approval-Not applicable
Consent to participate-Not applicable
Consent for publication-Not applicable
Availability of data and material-Proof of concept: https://review-detector-analysis.streamlit.app/; Data: https://github.com/alingabriel743/review-detector/blob/main/data/dataset.csv
Authors' contributions-A.V, S.V.O: Conceptualization, Methodology, Formal analysis, Investigation, Writing–Original Draft, Writing–Review and Editing, Visualization, Project administration. A.V, A.B: Validation, Formal analysis, Investigation, Resources, Data Curation, Writing–Original Draft, Writing–Review and Editing, Visualization, Supervision.

# Appendix A
## Prompt A. Adversarial review generation prompt template

```
You are a real human peer reviewer at a top ML conference. You have been reviewing papers for
10+ years. Write a genuine, natural peer review for this paper.

IMPORTANT STYLE REQUIREMENTS — your review must feel authentically human:
- Do NOT use markdown headers like "### Summary" or "### Strengths". Write in flowing
paragraphs or use simple formatting.
- Include personal voice: "I think", "I found this confusing", "after reading this twice", "in
my experience"
- Reference specific parts: mention page numbers, figure numbers, equation numbers, line
numbers, table numbers
- Be inconsistent in tone — mix formal and informal language naturally
- Show genuine uncertainty: "I might be wrong but...", "I'm not entirely sure about..."
- Vary your sentence length — mix short punchy sentences with longer ones
- Include minor imperfections: slight tangents, self-corrections, strong opinions
- Do NOT use balanced "Strengths/Weaknesses" lists. Real reviewers often focus more on one
side
- Occasionally be blunt or even a bit harsh — real reviewers sometimes are
- Reference your own expertise or related work you've read

Title: {title}

Abstract: {abstract}

Content: {content}

Write your review now. Make it 200-500 words, as a real reviewer would.
```

## Prompt B. Feature extraction prompt template

```
You are a linguistic analyst evaluating the writing characteristics of an academic peer
review.

Score each of the following 8 textual properties from 0.0 (not present at all) to 1.0 (very
strongly present). Be precise and use the full range of scores.

PROPERTIES:
1. standardized_structure: How rigidly does the text follow a templated structure with clearly
labeled sections (e.g., Summary, Strengths, Weaknesses)?
2. predictable_criticism: How much does the text rely on common, formulaic critique phrases
(e.g., "needs ablation study", "stronger baselines") rather than paper-specific criticism?
3. excessive_balance: How diplomatically balanced is the tone? Does it systematically pair
criticism with positive framing?
4. linguistic_homogeneity: How uniform are the grammar, sentence length, and tone throughout
the text?
5. generic_domain_language: How much does the text use broad academic phrases (e.g., "novel
approach", "significant contribution") rather than precise technical language?
6. conceptual_feedback: How much does the feedback stay at a high/conceptual level without
referencing specific lines, pages, figures, or tables?
7. absence_personal_signals: How absent are personal voice markers (e.g., "I think", "I
found", "in my experience", expressions of uncertainty)?
8. repetition_patterns: How much repetitive or templated phrasing appears across sections?

PEER REVIEW TEXT:
"""
{review_text}
"""

Respond ONLY with valid JSON containing the 8 scores (no justifications):
{
  "standardized_structure": 0.0,
  "predictable_criticism": 0.0,
  "excessive_balance": 0.0,
  "linguistic_homogeneity": 0.0,
  "generic_domain_language": 0.0,
  "conceptual_feedback": 0.0,
  "absence_personal_signals": 0.0,
  "repetition_patterns": 0.0
}
```